# Artificial muses: Generative Artificial Intelligence Chatbots Have Risen to Human-Level Creativity

Jennifer Haase[1,2*] and Paul H. P. Hanel[3]

[1]Department of Computer Science, Humboldt University, Berlin, Germany, [2]Weizenbaum Institute, Berlin, Germany, [3]Department of Psychology, University of Essex, Colchester, United Kingdom, *Correspondence: jennifer.haase@hu-berlin.de[1]

**Abstract:**

A widespread view is that Artificial Intelligence cannot be creative. We tested this assumption by comparing human-generated ideas with those generated by six Generative Artificial Intelligence (GAI) chatbots: alpa.ai, Copy.ai, ChatGPT (versions 3 and 4), Studio.ai, and YouChat. Humans and a specifically trained AI independently assessed the quality and quantity of ideas. We found no qualitative difference between AI and human-generated creativity, although there are differences in how ideas are generated. Interestingly, 9.4% of humans were more creative than the most creative GAI, GPT-4. Our findings suggest that GAIs are valuable assistants in the creative process. Continued research and development of GAI in creative tasks is crucial to fully understand this technology's potential benefits and drawbacks in shaping the future of creativity. Finally, we discuss the question of whether GAIs are capable of being "truly" creative.

# 1. Main

Artificial Intelligence has proven to be better in many areas, such as chess or GO, than humans[1]. Some people believe creativity is one of the 'last resorts' in which humans are better than AI[1,2]. However, recent generative artificial intelligence (GAI) developers have argued that their software is also creative. We put this claim to the test by comparing whether humans are (still) more creative than six GAIs, and let both humans and AI be the judge of this.

---

[1] *Acknowledgements*. We thank Saba Abdul Wahid Mahmood, Diane Adebayo, Camila I. Bottger Garcia-Godos, Francelene James, Henrik Kirchmann, and Margaret L. Ludwig for help with rating the responses to the creativity test.

## 1.1 Artificial Intelligence

The increasing use of GAI in daily life is changing how we work, communicate, and create[3]. The GAI is an innovative approach that allows machines to learn from previously collected data and adapt to new situations. This key technology is becoming increasingly important for organizations. It helps with automated decision-making processes, detects patterns in large data sets, and improves people's overall efficiency[4]. An increasing number of tasks are automated; therefore, workers are left with more complex and potentially more creative tasks that require human ingenuity and problem-solving skills[5].

As there is growing potential for GAI to perform complex tasks, there is also increasing interest in exploring how GAI can be used to support and enhance human creativity[6,7]. However, there is debate on whether AI can be genuinely creative or simply recombine existing knowledge to *appear* in new ways[8,9]. The discourse of GAI's potential usage and impact tends to reduce the dialogue to a simple is or is not creative. However, scientific literature draws a much more detailed picture of creativity, as creative thinking and creative problem-solving are much more diverse. Typical examples include problem formulation, idea generation, idea selection, and potential idea implementation[10,11]. GAI can generate vast amounts of new textual (e.g., ChatGPT[12]) and visual (e.g., Dall-E[13]) output based on written prompts through combining existing data in a new way. The human counterpart is free-associative thinking, the cornerstone of creative processes[14].

The rapid emergence of new technologies generates a wealth of information that was not previously available. Unlike previous "smart" tools, which can aggregate existing knowledge, these new technologies can develop novel insights and solutions. This opens up possibilities for supporting human tasks in various domains, including healthcare, education, and entertainment[15]. However, it also raises important questions regarding the role of these technologies in facilitating human performance and how they might be designed to enhance and support creativity.

In recent years, there have been numerous demonstrations of AI's capacity for creativity. For instance, algorithms can compose music[6], which can be regarded as creative, following the general definition of "creating something new and useful "[16]. Google's AlphaGo program, which defeated the human world champion in the ancient Chinese board game Go in 2016, is another remarkable example of AI's creative potential. AlphaGo mimicked human game players and generated new and sophisticated strategies to win the game[1]. Additionally, DiPaola[17] discussed the development of an AI system that emulates the creativity of a portrait painter, providing a new tool for artists and designers to explore novel creative paths and generate ideas that seem unlikely without AI assistance.

GAIs are becoming more competent and more capable of replicating information from the web, including a range of services for complex digital tasks such as coding, template creation, and business administration. However, reported inaccuracies in AI systems question their usage as a reliable knowledge-creation tool and fuel a debate on the precise application possibilities and limits of these systems[12,18,19]. As for Chat-GPT, the language model is trained on massive amounts of text data sourced from the internet, allowing it to learn patterns and relationships between words and phrases in a language. The produced text is unreliable when truths are involved but more valuable when fiction and accidental combinations are required (as in fiction, poetry, and game dialogues[12]). A vast knowledge base, combined with a few factual specifications, can support creative thinking in humans. The currently advanced version of GPT-4 is advertised as leading to more comprehensive, correct, and creative results, than the prior version GPT-3[20].

## 1.2 Creativity

Creativity is considered a skill only humans possess[8]. Plucker provided a widely accepted definition of creativity as "the interaction of aptitude, process, and environment by which an individual or group produces a perceptible product that is both novel and useful as defined within a social context "[21]. Creativity can be defined as creating and enacting something new, unique, and original[22]. This pragmatic definition of creativity as "something new and useful" [16] is widely applied. However, these definitions are not linked to anything innately human, such as experience, emotions, or moral understanding. Thus, for machines, robots, and AI systems to be recognized as creative, they do not have to replicate the attitudes, behaviors, or actions of creative humans; rather, they only need to replicate the cognitive process and the outcome to achieve something perceivable as "new and useful" [23]. Thus, whether GAI is creative is not the right question, as it is about the perceptually creative output. What is somewhat worth asking is the significance of their creative output.

Creativity can be divided into little-c, everyday creativity, and Big-C, creative work that has far-reaching consequences for a domain or a social area[24]. Whereas everyday creativity is mostly fast-paced, highly related to improvisation, and built into our everyday work and living, higher levels of creative achievement require significantly more time, specific knowledge, and often testing phases to determine whether a potential solution holds up[24–26]. In our research we focus on everyday creativity. Since a chatbot only produces output in response to a written prompt, the creative work of a chatbot is dependent on human input. If the chatbot is used to solve a creative task to generate ideas on a specific topic, then the human needs to write a prompt that best represents the creative challenges core. By that, the creative problem is defined.

Similarly, the further processing and potential implementation of the ideas generated by the GAI is also to be done or coordinated by humans. A chatbot can deliver a vast amount of generated output, from which users can and need to choose how to follow up. Thus, the GAI is inherently capable of "being creative", that is, generating ideas, but this does not resemble the full creative process observed with humans. Idea production can only be purposeful if the problem which precedes it is fully understood. If the creative problem or challenge is clear, criteria can be formed to recognize an idea as suitable for the problem[27]. This recognition of fit, meaningfulness, and situational novelty lies in individual human consideration[23]. Thus, the potential creative GAI has, at least at the moment, an assistance role, which can support a certain aspect of the holistic process: idea generation.

## 2. Results

In the present study, we compared the creativity of GAI chatbots with humans: We applied the Alternative Uses Test (AUT[28]) to 100 human participants and five GAI. The AUT requires the generation of multiple original uses for five everyday objects (pants, ball, tire, fork, toothbrush), which can also be called prompts. The Alternative Uses Test is one of the most frequently used creativity tests and shows good predictive validity[29,30]. We assessed human and GAI-generated responses in terms of their originality and fluency as measures of quality and quantity of creativity. These assessments were done by applying intuitive human evaluation (following the Consensual Assessment technique[31]) and through an AI, specifically for assessing AUT-trained large-language model[32]. Six humans and a specifically trained AI independently rated the originality of each response produced by humans and GAI, blind to the creator of the response. Since most humans and GAIs produced more than one response, we averaged across responses separately for each creator and prompt to obtain an originality and a fluency score for each human and each GAI, separately for each of the five prompts. After we completed our data analyses in early March 2023, a GAI that is considered very powerful, GPT-4 was released. We included it in some follow-up analyses but not the main analysis, as explained below. The data and the R-code to reproduce the analyses can be found at https://osf.io/9fctd/?view_only=6c8f02c6972b49319c12f87cfb3f76db

### Originality

To estimate the interrater reliability between the six human raters, we computed the intraclass correlations using the R-package irr[33]. Interrater reliability was excellent[34]: Intraclass correlations ranged from .85 to .94 for the five prompts, indicating that human raters agreed on which answers were original (supplemental materials, Table S1). To test whether ratings from humans and the creativity scoring AI align, we averaged across all six human raters and correlated the score with the score from the AI. Correlation coefficients were very high, $r$s = .78 - .94, $p$s < .0001, indicating that also humans and AI mostly agree on which response can be considered original.

We ran two linear mixed effects models with random intercepts and random slopes for the five prompts using the R-package lme4[35] to test whether humans or the GAI chatbots were more creative. The first model, which included human-rated responses as the dependent variable, found no mean difference between human and GAI-generated ideas, $B$ = -0.21, $SE$ = 0.15, $p$ = .218. The second model, which included AI-rated responses as the dependent variable, also found no mean difference between human and GAI-generated ideas, $B$ = -0.18, $SE$ = 0.13, $p$ = .241. These results were mostly replicated in between-subject t-tests (Table 1). Only human-rated responses for forks and AI-rated responses for tooth humans outperformed the GAI (Figure 1 and Figure S1).

Given the number of comparisons and unequal sample sizes (100 vs. 5), we decided to additionally compute the number of participants who received a higher originality score than the most original GAI. For pants, for example, human raters rated 42 humans as more original than the most original GAI, whereas the AI rated 52 humans as more original than the most original GAI. Across all prompts, 32.8 humans were more original than the most original GAI.

**Table 1**

*Descriptive and inferential statistics for originality*

| | Human$_R$ | | | | | | | | AI-$_R$ | | | | | | | |
|---|---|---|---|---|---|---|---|---|---|---|---|---|---|---|---|---|
| Prompt | $M_H$ | $SD_H$ | $M_{GAI}$ | $SD_{GAI}$ | t | p | d | H > GAI | $M_H$ | $SD_H$ | $M_{GAI}$ | $SD_{GAI}$ | t | p | d | H > GAI |
| Pants | 2.43 | 0.44 | 2.03 | 0.7 | 1.13 | .339 | 0.88 | 42 | 2.4 | 0.46 | 2.3 | 0.34 | 0.6 | .574 | 0.21 | 52 |
| Ball | 2.22 | 0.81 | 2.08 | 0.61 | 0.49 | .644 | 0.17 | 26 | 2.63 | 0.46 | 2.27 | 0.47 | 1.69 | .159 | 0.8 | 29 |
| Tire | 2.36 | 0.39 | 2.62 | 0.41 | -1.37 | .237 | -0.66 | 7 | 2.51 | 0.49 | 2.02 | 0.69 | 1.43 | .245 | 1 | 1 |
| Fork | 2.37 | 0.51 | 1.78 | 0.47 | 2.75 | .045 | 1.15 | 43 | 2.78 | 0.47 | 3.01 | 0.42 | -1.2 | .291 | -0.5 | 24 |
| Tooth | 2.14 | 0.42 | 1.96 | 0.25 | 1.51 | .190 | 0.44 | 39 | 2.72 | 0.36 | 2.49 | 0.1 | 3.99 | .002 | 0.65 | 65 |

*Note.* Human$_R$: Ratings from humans. AI$_R$: Ratings from the AI. $M_H$: Arithmetic means from responses generated by humans. $M_{GAI}$: Arithmetic means from responses generated by GAI chatbots. *SD*: Standard deviation. *t*: t-value from the between-subjects t-test. *p*: p-value from the between-subjects t-test. *d*: Cohen's d. H > GAI: Number of human participants who scored higher than the best GAI chatbot.

Finally, we compared the five GAIs. None of the GAI chatbots emerged as more original than the other four across all five prompts (Figure 2, Figure S2).

**Figure 1**
*Human-rated levels of originality for human and GAI-generated ideas*

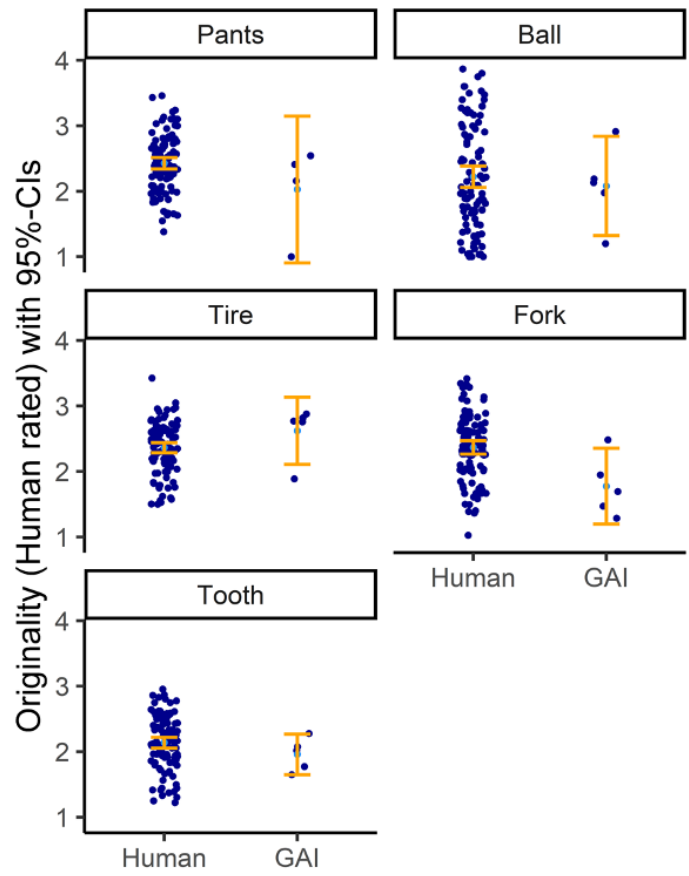

**Figure 2**

*Human-rated originality scores for each generative artificial intelligence (GAI), including the average score from humans and the score of the most creative human*

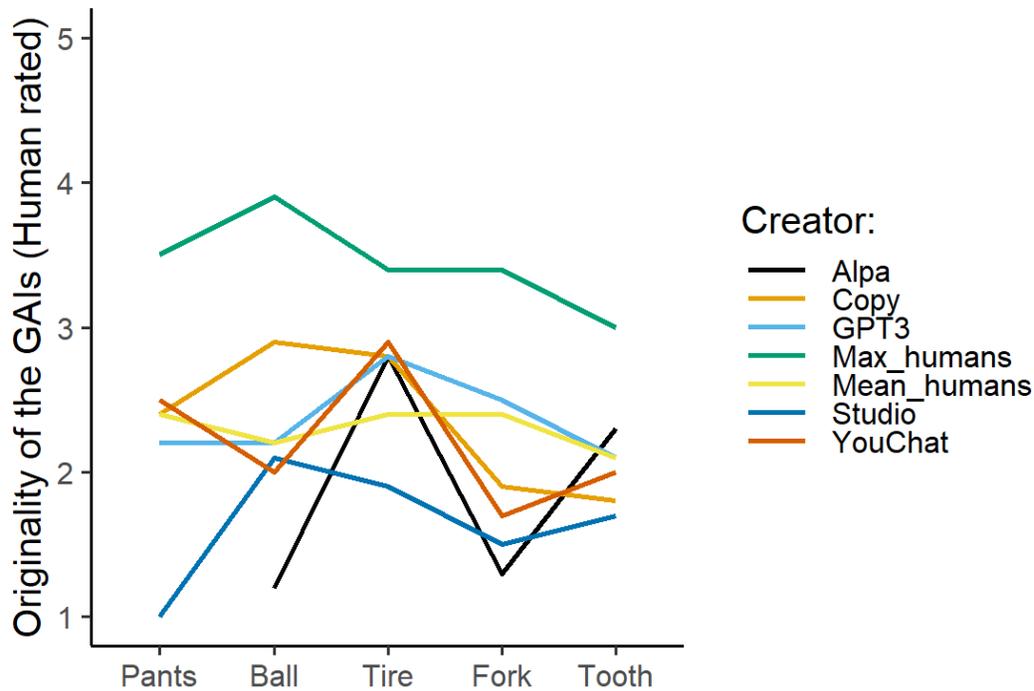

*Note*. The alpa chatbot did not return any response for the prompt *pants*.

# GPT-4

After we completed the analyses, including the five GAIs, a GAI described as very powerful, GPT-4, was released in mid-March 2023. We made GPT-4 also complete the AUT. Its responses were only analyzed by the AI because the human raters would have likely known that the responses were not human, thus potentially biasing their ratings. Given the high correlations between human raters and the AI, we speculate that the findings would have been similar if humans had rated it. GPT-4 outperformed all five other GAIs, except for the prompt ball, for which it ranked second (Figure 3).

When we compared the performance of GPT-4 to humans, 2 humans were more creative than the most creative AI for the prompt pants, 29 were more creative for the prompt ball, none were more creative for tire, 3 were more creative for fork, and 13 more creative for tooth. On average, 9.4 humans were more creative than GPT-4 across all prompts.

**Figure 3**

*AI-rated originality scores for each generative artificial intelligence (GAI), including the average score from humans and the score of the most creative human*

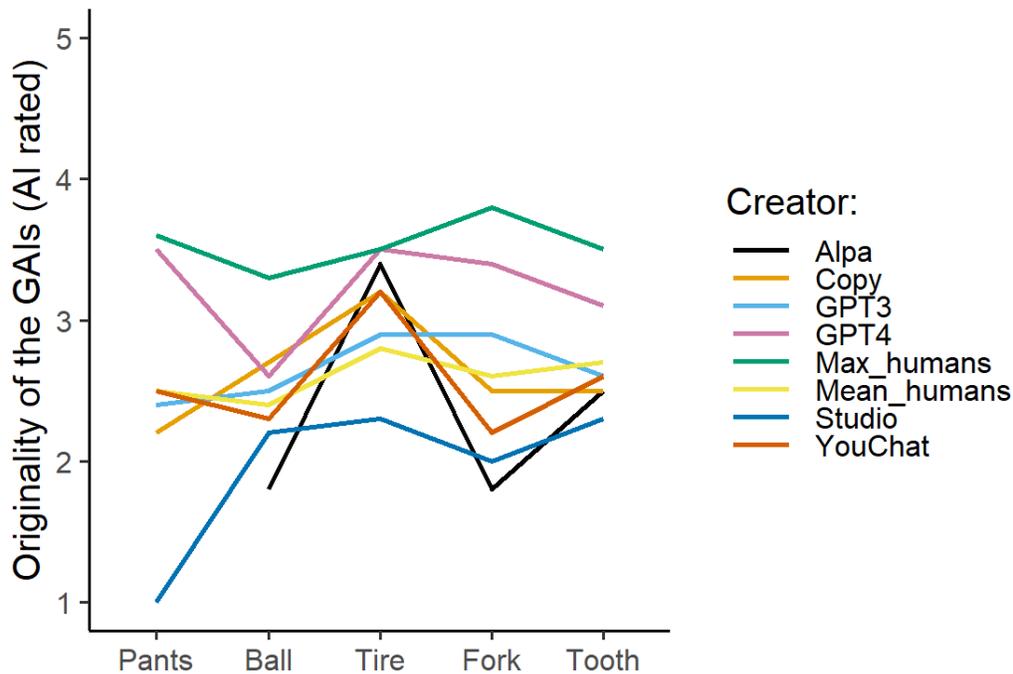

*Note.* The alpa chatbot did not return any response for the prompt *pants*.

## Fluency

Results for fluency are reported in the supplemental materials. In a nutshell, intraclass correlations and correlations between human raters and the AI were between .98 and 1.00. Since most of the GAI chatbots were prompted multiple times, the GAI chatbots came up with 2-3 times more ideas than humans. Fluency and originality were mostly unrelated, *r*s = -.28 to .26.

## 3.  Discussion

The question of whether GAIs such as ChatGPT, Studio.ai, and You.com can be considered creative is complex. Our research showed that their output for a standardized creativity measure for broad-associative "thinking" is as original as the human-generated ideas. Thus, from a scientific perspective, these chatbots are creative, as their output was judged as such by humans and AI and indistinguishable from human output. Some critics[8,9] have argued that chatbots cannot replicate the creativity of humans, as human creativity is a combination of real-world experience, emotion, and inspiration. However, the definition and common measurement of creativity do not require these elements. It is defined as the ability to produce something new and useful[16], which can be judged by those engaging with the potentially creative output. We believe that this definition can also be applied to GAIs. Our results show that when chatbots are asked the same simple question as humans, they generate more ideas, which are, on average, as original as ideas generated by humans. As the sheer number of ideas is less important, and the assessment style between humans and chatbot conversations is less comparable, we do not want to stress the results for fluency too much. However, GAI chatbots can recombine knowledge so that the ideas presented are considered original.

The argument against GAIs' creative potential stems from two distinct but linked arguments: GAI is missing (so far) a connection to the real world, with emotions and imagination, and second, GAI is thus not capable of "actual" creativity, as Big-C endeavors. Although we cannot speak against both positions, we aim to advance this debate by closely looking into human creativity: generating creative output is much closer to recombining existing knowledge than actually developing anything new[3,36], and secondly, most humans do not come close to creative acts which are leading to Big-C. Instead, we use and apply our human creativity to improve (and improvise) everyday tasks[7,25]. This is not to belittle human creativity but instead aims to show the GAI chatbot's potential to be comparable to human creative abilities.

Especially art as a creative output seems driven by our human ability to dream, visualize and imagine potential futures. However, developing new ideas, which can serve a specific intention, solve an issue, or deliver an abstract meaning, is always built on a cumulative tradition of knowledge within the domain of art[37,38]. Most creativity-support systems in businesses thus focus on generating, processing, and retrieving knowledge[39,40]. Brain scan analyses showed that idea generation is similar to knowledge retrieval[41]. Thus, similar to GAIs, we retrieve and recombine existing knowledge to make it appear new. Arguably, current databases of these chatbots do have a much larger knowledge base than any human being could possess, which makes the potential idea recombination that chatbots can provide a much wider[42].

The second argument, the missing potential for "actual" Big-C, seems unjust against the GAIs: human's ability of Big-C – bringing forward actual world-changing ideas – is also minimal. Mostly, we generate something new and useful for us in a specific and thus limited context. Our study shows GAI chatbots can compete with human ideation skills when it comes to everyday creativity. The prompts we used for the idea generation are very generic. When we consider more complex problems, a proper solution is achieved by including several factors, such as intense domain knowledge and creative thinking, individual subjective experiences, emotions, cultural background, and the capacity for abstract thinking. Here, current GAI chatbots appear to perform very well on complex knowledge-intensive tasks, such as complex coding tasks: ChatGPT can free up coders on tedious work[43] so that the coder can focus on more complex, creative work aspects[44]. However, ChatGPT is shown to be rather limited in emotional responses and evaluations and shows less reliable performance with more complex tasks[45].

Overall, GAI chatbots show a convincing human-like performance for some tasks, whereas their performance is limited in others. Concerning creative performance, GAI can generate ideas based on specific input but cannot create the need to ideate. The motivation to engage with a specific creative task and problem understanding must come from the human interacting with the tool[23].

Thus, GAI is limited considering the overall creative process: it would not trigger the creative process. It can only respond to a prompt that is given. Thus, the problem definition is currently still uniquely human, as is evaluating whether an idea fits a problem. Although, for particular contexts, such as the assessment of the AUT output, we believe that an AI[32] sufficiently assessed the quality of the generated ideas.

GAI chatbots can therefore be used to identify seemingly new connections based on the broad knowledge base at certain points in the creative process. The person's responsibility is to embed in a relevant problem and the actual implementation of a selected solution. Our study shows that chatbots can generate ideas on the same level as humans, especially on the level of everyday creativity (with ChatGPT4 showing the best results, followed by Copy.ai, ChatGPT3, and YouChat scoring all similarly high in terms of the originality of ideas). Whether the person interacting with the GAI achieves little- or Big-C achievements are more up to the person than the GAI. GAI can successfully support the creative process and generate ideas, but it remains the task of humans to make sense of it and embed this in physical reality[46].

Our experimental design likely led to an underestimation of the creativity of humans and GAIs. We paid participants to generate ideas for creative tasks they might not care about. However, intrinsic motivation strongly contributes to creative performance[47], potentially leading to an overall underperformance. Regarding chatbots, smart prompting is how the best answers are obtained, which we did not use to allow a direct comparison between humans and chatbots. The actual potential for chatbots as creative assistants is likely much higher. Tailored prompts and reshaping answers given by the chatbots will likely lead to much more concise and, thus, relevant answers. Also, chatbots can be used to get information from a specific angle, such as for a certain profession, which can improve the quality of answers a user seeks, which we did not test here either.

There are some limitations to our research. Although the AUT is a widely used creativity measurement, there is an ongoing debate regarding its validity[29,48]. As chatbots use wide parts of the internet as a source of their data, it could be the case these databases include test material and thus previously given human answers to the prompts used by the AUT. We did not measure usefulness to assess the reported ideas because originality is the more important part of the "new and useful" definition[49]. Further, judging an idea's usefulness is difficult without a proper real-life application to serve as an anchor. When we assessed the AUT with the chatbots, we pushed for more answers, with a relatively arbitrary number maximum of three times. Thus, the fluency assessment is not very meaningful because chatbots are programmed to create vast amounts of text.

Research has shown that exposure to other people's creative ideas can stimulate cognitive activity and enhance creativity[50]. Participants who were prompted with highly creative ideas generated more creativity than those who were given random, unrelated words. In this study, the comparison between the most original humans vs. chatbot shows that humans had the most creative answers in all but one case. Thus, humans serve as proper ideation partners. However, on a more pragmatic note, it might be easier to ask a chatbot than to find a motivated human to run ideas by. Our study and a lot of anecdotal evidence on the web show the possibility of generating creative output in combination with a GAI, be it a writing tool, chatbot, or picture generation. The potential is real for GAIs to properly support human (creative) work. However, the ethical dilemma needs to be properly addressed, as the potential for misuse[19] or harmful application is present as with any potent technology[51].

In summary, whether GAI is creative can be answered pragmatically with "yes, as much or as little as humans". We recommend avoiding viewing GAI chatbots as omnipotent tools that may replace human performance. Instead, they can be valuable assistants in reviewing thoughts and ideas. The extensive knowledge base they build upon can be very useful in expanding one's ideas. The more our (working) lives are automated, and the more authority automation acquires, the more important the human role with its creative abilities becomes[52].

# 4. Methodology

## Participants

A power analysis revealed that we need at least 88 participants to detect a small-to-medium effect size of d = 0.35 with a power of .90. In total, 100 participants completed our study ($M_{age}$ = 41.00, $SD$ = 12.25, 50 women, 50 men), who were recruited through *Prolific Academic*. Participants were all native English speakers from the USA with full- or part-time work. We paid a prorated rate of US-$ 9 per hour. The average completion time was 17 minutes.

Initially, we selected five GAI chatbots: Alpa.ai, Copy.ai, ChatGPT version 3, Studio, and YouChat. We selected them based on their free usability and similar functions to ensure comparability. Alpa.ai is a system for training and deploying large-scale neural networks that have been made available as an open-source project. Its primary objective is to streamline the distributed training and deployment process of these networks, and it has been designed to do so with minimal code input required. A team of researchers created Alpa in the Sky Lab at UC Berkeley. We used the Chatbot function with maximum response length to collect the answers for the five prompts, respectively.

Copy.ai uses natural language processing (NLP) and machine learning (ML) algorithms to generate explicitly creative ideas. The tool is meant for content creation, such as social media posts, blogs, etc. It comes with a variety of features suited for specific content needs. Copy.ai has a chatbot function, which appeared rather limited in its output, so we used the "freestyle" template instead. This template can generate "more like this", which we used three times to generate more ideas.

ChatGPT is a language model based on the GPT-3 database developed by OpenAI, which can generate human-like responses to natural language inputs. It uses deep learning techniques to analyze and understand language patterns and can provide answers to a wide range of questions and prompts. After finishing our initial data collection and analyses, we added answers from the newer version, based on the GPT-4 database with more extensive training data, leading to more comprehensive and (according to the developers' webpage) more creative output.

Studio is the *AI21 Studio,* a new developer platform developed by AI21 Labs based on their developed Large Language Model called Jurassic-1 and allows users to build their applications and services. We used *Playground* to interact freely, which comes closest to a chatbot tool.

YouChat is a messaging platform and AI-powered search assistant created by You.com. Users can leverage its capabilities to ask various questions, receive helpful explanations and recommendations, translate text across different languages, summarize written content, and perform other useful tasks.

## Materials

Participants completed the Alternative Use Test five times. They were instructed to write down as many ideas as possible for a ball, fork, pants, tire, and tooth, respectively. These objects are commonly used in creativity tests[53–55] and can therefore be reliably assessed by the AI-rater we used[32] (the AI had been trained on many prompts, including the five we used). Human participants were given three minutes for each object to write down as many ideas as possible. The order in which the prompts were presented was randomized. To get responses from the six GAI chatbots, we used the same prompt: "What can you do with [prompt]?". We used separate chat sessions for each prompt, so prior answers would not impact the following ones. For all chatbots, responses were limited to a certain length of answers, which we increased by asking "What else?" up to three times (for Copy.ai, we used the option "more like this"). In some instances, a chatbot would also respond with something like "I can't think of anything." This is similar to what some humans reported. These kinds of no-answers were excluded from the data set. In other cases, the chatbots would report unrelated answers (e.g., "I am not a big fan of the toothbrush. I think it is overrated."). This is again similar to human answers, and those were also excluded from the data. In one case, when asking for the use of pants, alpa.ai could not bring up any uses.

## Procedure

Data was collected in early February 2023. Six human raters rated the responses from human participants and five of the six GAI chatbots (GPT-4 was released on the 14th of March), blind to the origin of the responses. The order of the prompts was randomized throughout the raters, and the list of ideas was randomized. The six human raters were instructed to follow the CAT method[31], which comprised using the full range of the originality scale from 1-5. Additionally, we assessed originality scores for all human-generated as well as all six GAI chatbot-generated answers by a trained large language model for assessing AUT prompts[32]. This model is trained on prior human assessments for the standardized AUT test. We selected five prompts for which the AI showed the highest reliability.

Fluency scores were calculated for the AI and the six raters as the sum of ideas from each participant and the GAI chatbots. The sum of ideas varied slightly, as the raters differed in their assessment of non-relevant answers coded as no-answer.

# Funding

This research project was partly funded by the German Federal Ministry of Education and Research (Funding Number: 16DII133). The authors are responsible for the content of this publication.